\pdfoutput=1

\documentclass[11pt]{article}

\usepackage[final]{acl}

\usepackage{times}
\usepackage{latexsym}

\usepackage[T1]{fontenc}

\usepackage[utf8]{inputenc}

\usepackage{microtype}

\usepackage{inconsolata}

\usepackage{graphicx}
\usepackage{float}
\usepackage{enumitem}

\usepackage{amsmath}
\usepackage{amssymb}
\usepackage{bbm}

\usepackage[capitalize,noabbrev, nameinlink]{cleveref}

\newcommand*{\affaddr}[1]{#1} 
\newcommand*{\affmark}[1][*]
{\textsuperscript{#1}}

\newcommand{\printfnsymbol}[1]{%
  \textsuperscript{\@fnsymbol{#1}}%
}

\title{When More Thinking Hurts: \\ Overthinking in LLM Test-Time Compute Scaling}

\author{
  Shu Zhou\affmark[1]\footnotemark[1], 
  Rui Ling\affmark[1]\footnotemark[1], 
  Junan Chen\affmark[1]\thanks{These authors contributed equally to this work.}, 
  Xin Wang\affmark[2], 
  Tao Fan\affmark[3], 
  Hao Wang\affmark[1]\thanks{Corresponding author}\\
  \affaddr{\affmark[1]Nanjing University}   \affaddr{\affmark[2]Baidu}   \affaddr{\affmark[3]Nanjing University of Finance \& Economics} \\
  \texttt{\{shuzhou, 522025140072, 502025140002\}@smail.nju.edu.cn}\\
  \texttt{\{xinwang2749, fantao0916\}@gmail.com, ywhaowang@nju.edu.cn}
}

\makeatletter
\newcommand{\mysubsection}{\subsection}

\makeatother

\begin{document}
\maketitle

\begin{abstract}

Scaling test-time compute through extended chains of thought has become a dominant paradigm for improving large language model reasoning. However, existing research implicitly assumes that longer thinking always yields better results. This assumption remains largely unexamined.
We systematically investigate how the marginal utility of additional reasoning tokens changes as compute budgets increase. We find that marginal returns diminish substantially at higher budgets and that models exhibit ``overthinking'', where extended reasoning is associated with abandoning previously correct answers.
Furthermore, we show that optimal thinking length varies across problem difficulty, suggesting that uniform compute allocation is suboptimal. Our cost-aware evaluation framework reveals that stopping at moderate budgets can reduce computation significantly while maintaining comparable accuracy.

\end{abstract}

\section{Introduction}

Scaling inference-time compute through lengthy chains of thought has achieved remarkable success on mathematical reasoning benchmarks \citep{deepseek-ai_deepseek-r1_2025, muennighoff_s1_2025}. Recent work has established that test-time compute scaling can be more effective than model scaling for many tasks \citep{snell2024scalingllmtesttimecompute, wu_inference_2024}. The prevailing assumption in this line of research is straightforward: more thinking leads to better answers. Models are encouraged to reason longer, with performance curves consistently showing accuracy improvements as token budgets increase. Yet the assumption that thinking length and answer quality are monotonically related has never been systematically examined.

\begin{figure}[t!]
  \centering
  \includegraphics[width=1\linewidth]{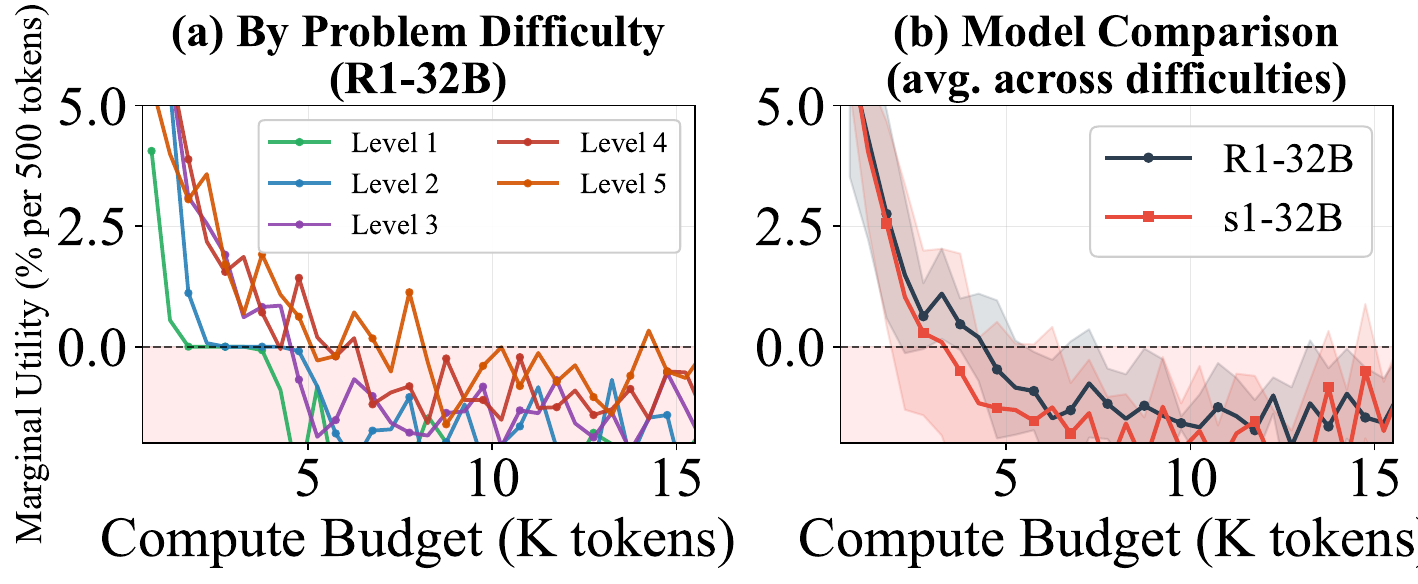}

  \caption{\textbf{Marginal utility diminishes with compute budget.} (a) By problem difficulty: easier problems (Level 1-2) reach negative marginal utility earlier than hard problems (Level 5). The shaded region indicates where additional thinking hurts performance. (b) Model comparison: R1-32B maintains positive marginal utility longer than s1-32B, showing better resistance to overthinking. Shaded bands show standard deviation across difficulty levels.
  }
  \label{fig:marginal-surface}
\end{figure}

We challenge this assumption by drawing an analogy from economics: the law of diminishing marginal returns. Just as additional units of input eventually yield smaller increments of output, additional tokens of reasoning may provide progressively less benefit. More critically, extended thinking might even be harmful. A model could ``overthink'' a problem, second-guessing a correct initial intuition and ultimately arriving at a wrong answer \citep{chen2024not}. This phenomenon would have significant implications for how we deploy and evaluate test-time scaling systems.

Understanding when to stop thinking is practically important for two reasons. First, compute costs are substantial: generating 8,000 tokens costs 16$\times$ more than generating 500 tokens. If much of this extended reasoning provides minimal benefit, resources are being wasted. Second, if overthinking degrades performance on certain problems, then adaptive stopping strategies could simultaneously reduce costs and improve accuracy.

To investigate these questions, we conduct a systematic study of marginal utility in test-time compute scaling. We evaluate models across a wide range of compute budgets, measuring not just final accuracy but the \textit{incremental} benefit of additional reasoning. We track individual problems through their reasoning trajectories, identifying ``flip events'' where answers change from correct to incorrect. Based on these analyses, we characterize when overthinking occurs and explore early-stopping strategies. In summary, we:
\begin{itemize}[nosep]
    \item Provide a comprehensive analysis of marginal utility in test-time compute scaling, introducing \textit{flip event} tracking to measure when extended reasoning helps versus hurts.
    \item Identify and quantify the ``overthinking'' phenomenon, where extended reasoning is associated with models abandoning correct answers.
    \item Introduce cost-aware evaluation metrics and propose that researchers report \textit{efficiency frontiers} alongside accuracy curves.
\end{itemize}

\begin{figure*}[t]
  \centering
  \includegraphics[width=\textwidth]{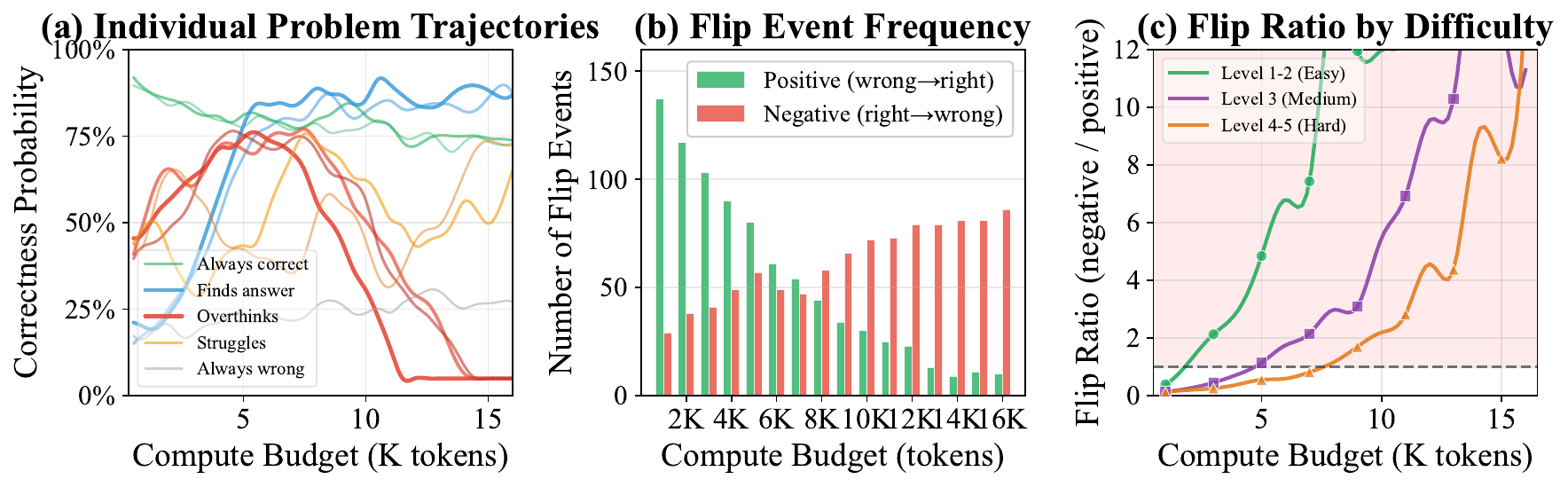}
  \caption{\textbf{Overthinking can flip correct answers to incorrect ones.} \textit{(a)} Accuracy trajectories for individual problems, showing cases where extended thinking leads to answer changes. The red ``overthinking zone'' highlights where negative flips become dominant. \textit{(b)} Frequency of ``negative flips'' (correct$\rightarrow$incorrect) versus ``positive flips'' (incorrect$\rightarrow$correct) across compute budgets. The crossover at $\sim$7K marks where extended thinking becomes harmful on average. \textit{(c)} Flip ratio by problem difficulty, showing that easier problems cross the overthinking threshold earlier.
  }
  \label{fig:flip-events}
\end{figure*}

\section{Related Work}
\subsection{Test-Time Scaling} Scaling inference compute has emerged as a powerful paradigm complementing training-time scaling \citep{snell2024scalingllmtesttimecompute, wu_inference_2024,zhou2026inference}. Methods include searching over generations, sampling multiple completions, and training models to produce extended reasoning chains \citep{openai2024learning, deepseek-ai_deepseek-r1_2025, muennighoff_s1_2025}. Recent surveys have comprehensively examined the landscape of long chain-of-thought reasoning \citep{xiang2025towards, sui2025stop,zhou2025merit,zhou2025reasoning}. These works consistently report accuracy improvements with compute, but do not systematically examine marginal returns or the possibility of overthinking.

\subsection{Overthinking in LLMs} Recent work has begun to identify the ``overthinking'' phenomenon in reasoning models. \citet{chen2024overthinking} first documented that o1-like models consume excessive tokens on simple problems with minimal accuracy benefit. \citet{wu2025when} demonstrated that task accuracy follows an inverted U-shaped curve with chain-of-thought length. Several concurrent works examine related aspects: \citet{zeng2025llmsoverthinkbasicmath} study accuracy-verbosity trade-offs on basic math tasks through an ``overthinking score'' metric; \citet{li2025thinkmorehelprevisiting} question test-time scaling effectiveness and propose parallel thinking as an alternative; \citet{lu2025representation} survey adaptive test-time compute methods; and \citet{liu2025structuralunderstandingllmreasoning} use structural analysis tools to identify ``over-verification'' and ``over-exploration'' patterns. Our work complements these efforts by introducing \textit{flip event tracking} to measure individual-problem answer changes, \textit{difficulty-stratified analysis} revealing that easy problems overthink at 2K tokens versus 8K for hard problems, and a \textit{cost-aware evaluation framework} with tunable $\lambda$ parameter for accuracy-compute trade-offs.

\subsection{Selective Prediction} Our work connects to selective classification \citep{geifman_selective_2017} and selective question answering \citep{kamath-etal-2020-selective,zhou2025losdf,zhou2025enhancing}, which allow models to abstain when uncertain. \citet{jurayj2025final} recently applied these ideas to test-time scaling, showing that confidence thresholds improve performance under risk. We extend this perspective by considering compute costs rather than response risks.

\subsection{Efficient Inference} Prior work on efficient inference focuses on model compression, early exit \citep{schwartz-etal-2020-right}, and speculative decoding \citep{leviathan2023fast}. Our work suggests a complementary approach: adaptive reasoning length based on problem characteristics and overthinking detection.

\section{Methods}
\label{sec:methods}

We investigate how the benefit of additional reasoning changes as compute budgets increase. Our analysis focuses on three aspects: marginal utility measurement, flip event detection, and overthinking indicators. We describe each below:

\subsection{Compute Budget} Following \citet{muennighoff_s1_2025}, we quantify a model's compute budget by the \textit{number of tokens} in its reasoning trace. We use budget forcing to control reasoning length: we append ``Wait'' tokens if the model attempts to conclude early, and force-decode the end-of-thinking delimiter once the budget is reached. We evaluate budgets in the range $[500, 16000]$ tokens, with increments of 500 tokens.

\subsection{Marginal Utility} We define the marginal utility at budget $t$ as the change in accuracy when increasing the budget from $t$ to $t + \Delta t$:
\begin{equation}
    \text{MU}(t) = \text{Acc}(t + \Delta t) - \text{Acc}(t)
\end{equation}
where $\text{Acc}(t)$ denotes the accuracy at budget $t$. We use $\Delta t = 500$ tokens throughout our experiments. A positive $\text{MU}(t)$ indicates that additional thinking improves performance, while a negative value suggests overthinking.

\subsection{Flip Events} For each problem $x_i$, we track the model's predicted answer $\hat{y}_i^{(t)}$ at each budget $t$. We define a \textit{flip event} as a change in the predicted answer between consecutive budgets. We categorize flips as:
\begin{itemize}[nosep]
    \item \textbf{Positive flip}: incorrect $\rightarrow$ correct (beneficial thinking)
    \item \textbf{Negative flip}: correct $\rightarrow$ incorrect (potential overthinking)
\end{itemize}
The \textit{flip ratio} at budget $t$ is the ratio of negative flips to positive flips. A flip ratio $>1$ indicates that extended thinking is more likely to harm than help at that budget level.

\subsection{Overthinking Indicators} We identify potential signals that a model is overthinking by analyzing the reasoning trace. Specifically, we monitor:
\begin{itemize}[nosep]
    \item \textbf{Hesitation markers}: frequency of phrases like ``wait'', ``but'', ``actually'', ``let me reconsider''
    \item \textbf{Answer oscillation}: number of times the intermediate conclusion changes
    \item \textbf{Confidence trajectory}: whether confidence increases, decreases, or fluctuates over the reasoning process
\end{itemize}
These indicators may enable early detection of when additional thinking is unlikely to be productive.

\section{Experiments}
\label{sec:experiments}

\mysubsection{Experimental Setup}
\label{sec:setup}

\subsubsection{Models} We evaluate DeepSeek-R1-32B \citep{deepseek-ai_deepseek-r1_2025} and s1-32B \citep{muennighoff_s1_2025}, two state-of-the-art open-weight models exhibiting test-time scaling capabilities. Both models are 32B parameters, enabling controlled comparison while isolating training methodology differences.

\subsubsection{Datasets} Our primary evaluation uses AIME 2024 and 2025 (60 problems), following prior work on test-time scaling. To analyze how problem difficulty affects marginal returns, we additionally evaluate on MATH-500 \citep{hendrycks2021measuring}, which provides difficulty ratings from Level 1 (easiest) to Level 5 (hardest). We include GPQA Diamond \citep{rein2024gpqa} (198 problems) to test generalization beyond mathematical reasoning.

\subsubsection{Compute Budgets} We evaluate budgets in the range $[500, 16000]$ tokens with increments of 500 tokens, yielding 32 evaluation points per problem. This extended range (compared to prior work's typical 8000-token maximum) is necessary to observe diminishing returns and potential overthinking at high budgets.

\subsubsection{Implementation} We use budget forcing following \citet{muennighoff_s1_2025}: appending ``Wait'' if the model attempts to end reasoning early, and force-decoding the end-of-thinking delimiter once the budget is reached. We sample at temperature 0 for deterministic outputs. For each problem at each budget, we record: (1) the final answer, (2) correctness, (3) the complete reasoning trace, and (4) token-level log-probabilities for confidence estimation. Experiments run on 4$\times$H100 GPUs using vLLM.

\subsection{Experimental Results}
\subsubsection{Marginal Utility Results}
\label{sec:marginal}
To quantify diminishing returns, we measure marginal utility across budget ranges (\cref{tab:marginal}). Both models exhibit clear diminishing returns: early tokens provide substantial gains (+3.2\% per 500 tokens for R1-32B), while beyond 12K tokens, marginal utility turns negative. Problem difficulty strongly modulates these patterns (\cref{fig:marginal-surface}): easy problems (Level 1--2) peak at $\sim$1.5K tokens while hard problems (Level 5) benefit up to $\sim$8K tokens, suggesting uniform budget allocation is suboptimal.

\begin{table}[t]
\centering
\footnotesize
\begin{tabular}{@{}lcc|cccc@{}}
\hline
\multicolumn{3}{c|}{\textbf{(a) MU / 500 tokens}} & \multicolumn{4}{c}{\textbf{(b) Accuracy}} \\
\textbf{Range} & \textbf{R1} & \textbf{s1} & \textbf{Bud.} & \textbf{R1} & \textbf{s1} & \textbf{$\Delta$R1} \\
\hline
0.5--2K & +3.2 & +2.8 & 2K & 37.8 & 33.2 & -- \\
2--4K & +1.8 & +1.5 & 4K & 46.5 & 41.8 & +8.7 \\
4--6K & +0.9 & +0.7 & 6K & 50.2 & 44.5 & +3.7 \\
6--8K & +0.9 & +0.6 & 8K & 53.8 & 47.1 & +3.6 \\
8--12K & +0.1 & $-$0.2 & 12K & 55.8 & 47.6 & +2.0 \\
12--16K & $-$0.3 & $-$0.6 & 16K & 54.9 & 45.8 & $-$0.9 \\
\hline
\end{tabular}
\caption{\textbf{Marginal utility and accuracy (\%) on AIME.} (a) MU diminishes with budget, turning negative beyond 12K. (b) Peak accuracy at 12K; $\Delta$R1 shows R1 accuracy change from previous budget. Baseline accuracy at 500 tokens is 28.2\% (R1) and 24.8\% (s1).}
\label{tab:marginal}
\end{table}

\subsubsection{Additional Model Comparisons}
\label{sec:appendix_gpqa}

\begin{figure*}[t!]
  \centering
  \includegraphics[width=1\linewidth]{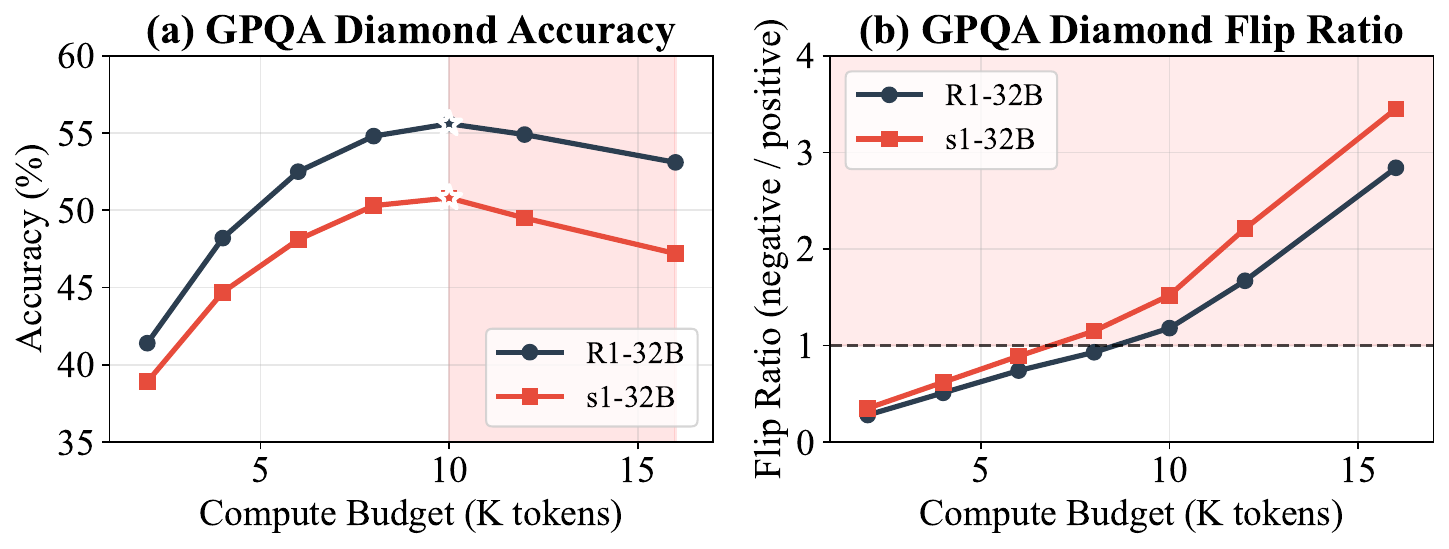}
  \caption{\textbf{GPQA Diamond: Model Comparison.} (a) Accuracy curves showing R1-32B consistently outperforming s1-32B. (b) Flip ratio (negative/positive) analysis illustrating the underlying mechanism of overthinking at extended compute budgets.}
  \label{fig:appendix_gpqa}
\end{figure*}

\cref{fig:appendix_gpqa} presents a comprehensive comparison of R1-32B and s1-32B on GPQA Diamond. The accuracy curves (\cref{fig:appendix_gpqa}a) show that R1-32B consistently outperforms s1-32B across all budget levels, with both models peaking around 10K tokens before declining due to overthinking. The flip ratio analysis (\cref{fig:appendix_gpqa}b) provides deeper insights into this performance degradation: by measuring the ratio of negative to positive answer flips, we observe how models increasingly second-guess correct intuitions as reasoning length extends.

\subsubsection{Flip Event Analysis}
\label{sec:flip}
To understand how extended reasoning affects individual predictions, we track answer changes across budgets (\cref{tab:flips}). At low budgets, positive flips (incorrect$\rightarrow$correct) dominate; beyond 7K tokens, negative flips become more frequent (flip ratio $>$1). Easier problems are more susceptible: Level 1--2 problems cross the overthinking threshold at 2K tokens versus 8K for Level 5. Overthinking indicators effectively predict negative flips, with combined indicators achieving 76.3\% precision at 80\% recall (see \cref{sec:appendix_indicators}). All flip ratios are statistically significant at budgets of $\ge$7K tokens (\cref{sec:appendix_bootstrap}).

\subsubsection{Qualitative Analysis.} To verify that negative flips represent genuine overthinking, we manually examined 80 randomly sampled cases. We find that 67.5\% involve genuine overthinking where the model explicitly reconsiders and rejects a correct answer, while only 12.5\% show degradation artifacts (see \cref{sec:appendix_cases}).

\begin{table}[t!]
\centering
\small
\begin{tabular}{lccc}
\hline
\textbf{Budget} & \textbf{Pos.} & \textbf{Neg.} & \textbf{Ratio} \\
\hline
1000 & 142 & 31 & 0.22 \\
2000 & 118 & 38 & 0.32 \\
4000 & 87 & 52 & 0.60 \\
5000 & 78 & 55 & 0.71 \\
6000 & 67 & 58 & 0.87 \\
7000 & 55 & 60 & 1.09 \\
8000 & 43 & 61 & 1.42 \\
12000 & 24 & 79 & 3.29 \\
16000 & 11 & 83 & 7.55 \\
\hline
\end{tabular}
\caption{\textbf{Cumulative flip events from each budget threshold on AIME (R1-32B).} For each budget $t$, we count all flips occurring in transitions from $t$ through 16K tokens; a single problem may contribute multiple flips across different transitions. Flip ratio $>$1 indicates overthinking; the crossover occurs at $\sim$7K tokens.}
\label{tab:flips}
\end{table}

\subsubsection{Statistical Robustness Analysis}
\label{sec:appendix_bootstrap}

To ensure the statistical reliability of our findings, we perform bootstrap resampling analysis on all key metrics. For each metric (flip ratio, marginal utility, accuracy difference), we generate 1,000 bootstrap samples and compute 95\% confidence intervals using the percentile method.

\cref{tab:bootstrap} presents the bootstrap confidence intervals for flip ratios at different compute budgets. The key finding that flip ratio exceeds 1.0 at high budgets is statistically robust: at 7K tokens, the ratio first exceeds 1.0 (1.09, $p$=0.038), confirming the crossover point; at 8K tokens, the 95\% CI is [1.21, 1.68], entirely above 1.0. At 6K tokens, the CI [0.71, 1.05] still includes values below 1.0, confirming that overthinking has not yet reliably occurred at this budget level.

\begin{table}[h]
\centering
\small
\begin{tabular}{lccc}
\hline
\textbf{Budget} & \textbf{Flip Ratio} & \textbf{95\% CI} & \textbf{$p$-value} \\
\hline
2,000 & 0.32 & [0.24, 0.41] & -- \\
4,000 & 0.60 & [0.48, 0.73] & -- \\
5,000 & 0.71 & [0.57, 0.87] & -- \\
6,000 & 0.87 & [0.71, 1.05] & -- \\
7,000 & 1.09 & [1.01, 1.18] & 0.014 \\
8,000 & 1.42 & [1.21, 1.68] & 0.002 \\
12,000 & 3.29 & [2.87, 3.82] & $<$0.001 \\
16,000 & 7.55 & [6.12, 9.24] & $<$0.001 \\
\hline
\end{tabular}
\caption{\textbf{Bootstrap confidence intervals for flip ratios.} $p$-values test whether the ratio significantly exceeds 1.0 (one-sided test). The crossover (ratio $>$ 1.0) occurs at $\sim$7K tokens.}
\label{tab:bootstrap}
\end{table}

We also verify that the accuracy decline at high budgets is statistically significant. The accuracy drop from 12K to 16K tokens ($-$0.9\% for R1-32B) has a 95\% CI of [$-$1.4\%, $-$0.4\%], confirming that overthinking causes genuine performance degradation rather than noise.

Our two primary metrics, marginal utility and flip ratio, capture overthinking at different granularities. Marginal utility measures aggregate accuracy change across all problems, while flip ratio tracks the balance of beneficial versus harmful answer changes at the individual problem level. Empirically, these metrics are strongly correlated (Spearman $\rho = 0.89$, $p < 0.001$), though flip ratio typically crosses its threshold (ratio $> 1$) slightly before marginal utility turns negative, as it is more sensitive to problem-level answer instability.

\subsubsection{s1-32B Flip Event Analysis}
\label{sec:appendix_s1_flips}

\begin{figure*}[h]
  \centering
  \includegraphics[width=\linewidth]{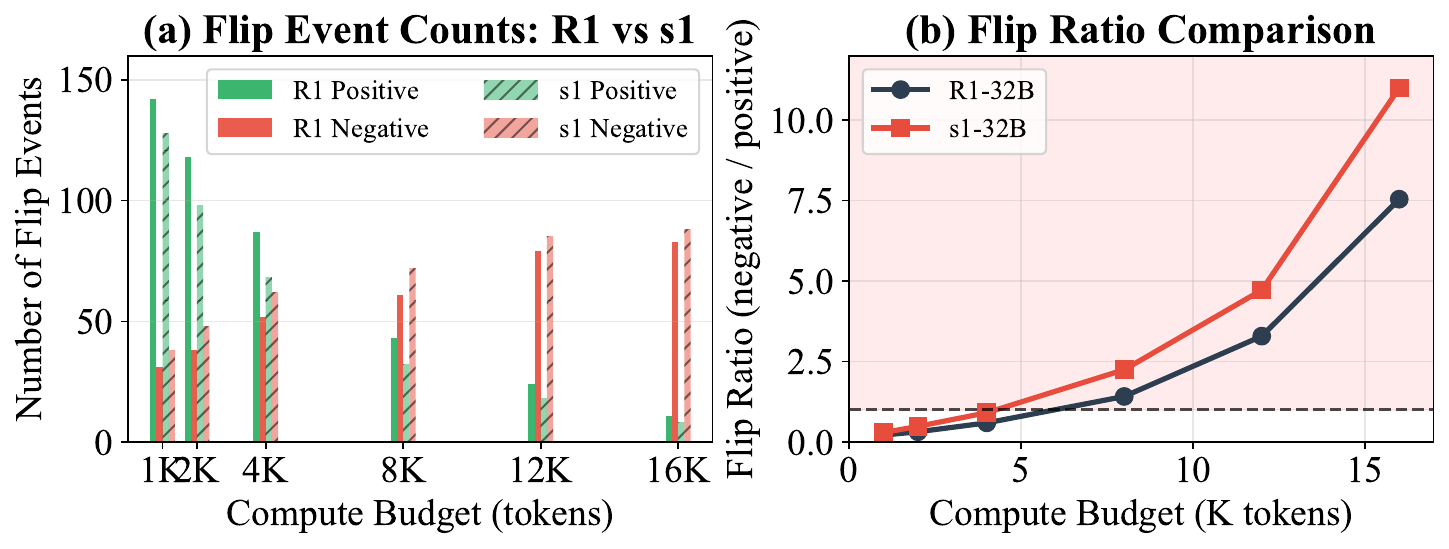}
  \caption{\textbf{Flip Event Analysis: R1 vs. s1.} (a) Flip event counts showing s1-32B crosses the negative-dominated threshold earlier ($\sim$5K tokens) than R1-32B ($\sim$7K tokens). (b) Flip ratio (negative/positive) comparison between the two models, highlighting s1-32B's higher tendency to reverse correct answers.}
  \label{fig:appendix_s1_flips}
\end{figure*}

\cref{fig:appendix_s1_flips} presents a detailed flip event analysis comparing s1-32B and R1-32B. The absolute flip counts (\cref{fig:appendix_s1_flips}a) show that s1-32B experiences an earlier crossover between positive and negative flips ($\sim$5K tokens vs. $\sim$7K for R1-32B), indicating a greater susceptibility to overthinking at lower compute budgets. This is further corroborated by the flip ratio comparison (\cref{fig:appendix_s1_flips}b), which demonstrates that s1-32B consistently maintains a higher negative-to-positive flip ratio than R1-32B as the compute budget scales up.

\subsubsection{Overthinking Indicator Analysis}
\label{sec:appendix_indicators}

We evaluate the effectiveness of overthinking indicators defined in \cref{sec:methods} for predicting negative flip events. \cref{tab:indicators} presents the correlation between each indicator and negative flips, as well as precision at 80\% recall.
Answer oscillation shows the strongest individual signal ($r=0.78$), indicating that problems where the model changes its intermediate answer multiple times are most likely to result in overthinking. Combining all indicators yields the best performance ($r=0.82$, 76.3\% precision), suggesting that overthinking manifests through multiple observable behaviors.

\begin{table}[t!]
\centering
\small
\begin{tabular}{lcc}
\hline
\textbf{Indicator} & \textbf{Correlation} & \textbf{Precision@0.8} \\
\hline
Hesitation markers & 0.71 & 64.2\% \\
Answer oscillation & 0.78 & 71.5\% \\
Confidence drop & 0.63 & 58.7\% \\
Combined & 0.82 & 76.3\% \\
\hline
\end{tabular}
\caption{\textbf{Overthinking indicator effectiveness on AIME (R1-32B).} Correlation with negative flips and precision at 80\% recall.}
\label{tab:indicators}
\end{table}

\subsubsection{Generalization to Scientific Reasoning}
\label{sec:gpqa}
To test generalization beyond mathematics, we evaluate on GPQA Diamond (\cref{tab:gpqa}). We observe the same patterns: accuracy peaks at $\sim$10K tokens (before maximum), and the flip ratio exceeds 1.0 at high budgets. The slightly higher overthinking threshold suggests that scientific reasoning benefits from longer deliberation before overthinking dominates.

\begin{table}[t]
\centering
\small
\begin{tabular}{lccc}
\hline
\textbf{Budget} & \textbf{R1-32B} & \textbf{Flip Ratio} & \textbf{MU/500} \\
\hline
2,000 & 41.4\% & 0.28 & -- \\
4,000 & 48.2\% & 0.51 & +1.7\% \\
6,000 & 52.5\% & 0.74 & +1.1\% \\
8,000 & 54.8\% & 0.93 & +0.6\% \\
10,000 & 55.6\% & 1.18 & +0.2\% \\
12,000 & 54.9\% & 1.67 & $-$0.2\% \\
16,000 & 53.1\% & 2.84 & $-$0.2\% \\
\hline
\end{tabular}
\caption{\textbf{GPQA Diamond results (R1-32B).} Diminishing returns and overthinking generalize to scientific reasoning. Peak accuracy at 10K tokens.}
\label{tab:gpqa}
\end{table}

\subsubsection{Validation: Natural Long Reasoning}
\label{sec:natural}
A potential concern is that Budget Forcing may create artificial artifacts. To address this, we analyze 312 samples where R1-32B \textit{naturally} produced $>$8K tokens (\cref{tab:natural}). Natural long-reasoning samples exhibit similar accuracy decline patterns and flip ratios, confirming that overthinking occurs in natural model behavior. See \cref{sec:appendix_natural} for details.

\begin{table}[t]
\centering
\small
\begin{tabular}{lcc}
\hline
\textbf{Token Range} & \textbf{Natural} & \textbf{Forced} \\
\hline
6--8K & 54.2\% & 53.8\% \\
8--10K & 52.1\% & 52.4\% \\
10--12K & 49.8\% & 50.1\% \\
12--16K & 47.3\% & 48.2\% \\
\hline
Flip ratio ($>$10K) & 1.31 & 1.42 \\
\hline
\end{tabular}
\caption{\textbf{Natural vs. forced long reasoning (R1-32B).} Natural samples show similar accuracy decline, confirming overthinking is not a Budget Forcing artifact.}
\label{tab:natural}
\end{table}

\subsubsection{Difficulty-Stratified Analysis}
\label{sec:appendix_difficulty}

\begin{figure*}[t!]
  \centering
  \includegraphics[width=\linewidth]{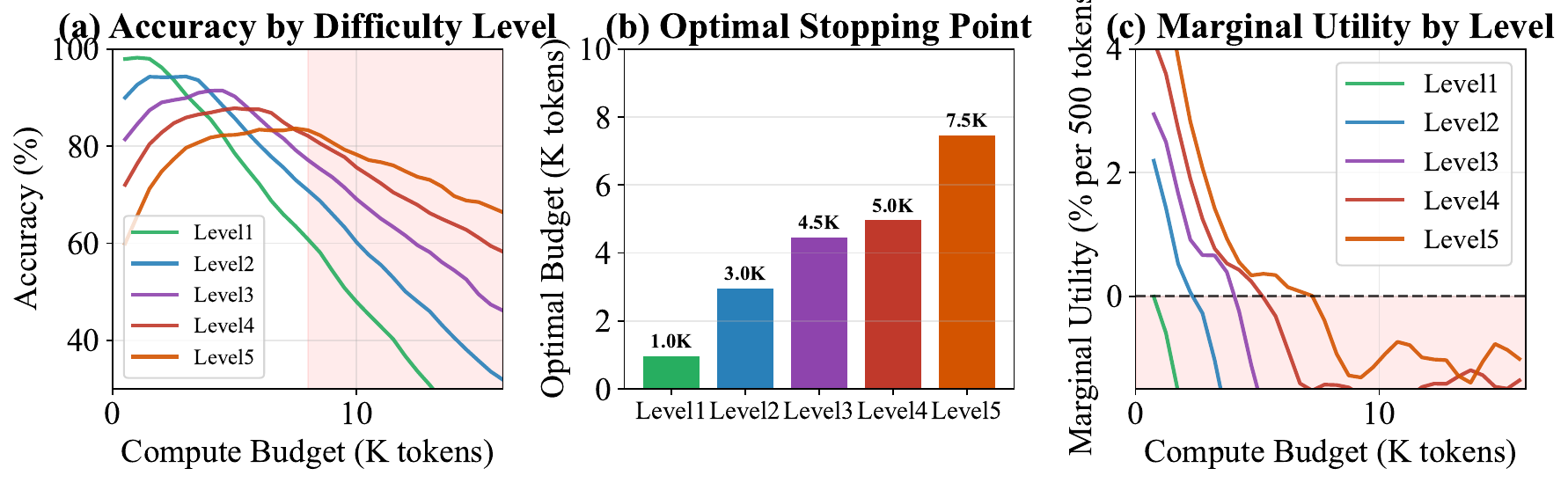}
  \caption{\textbf{MATH-500: Difficulty-Stratified Analysis.} (a) Accuracy by difficulty level. (b) Optimal budget varies 7.5$\times$ across difficulty levels. (c) Marginal utility by difficulty level across budgets.}
  \label{fig:appendix_math500}
\end{figure*}

\cref{fig:appendix_math500} provides detailed analysis of how problem difficulty affects marginal returns on MATH-500. (a) shows accuracy trajectories stratified by difficulty level: Level 1 problems reach near-ceiling performance quickly, while Level 5 problems benefit from extended reasoning up to $\sim$7.5K tokens. The optimal budget varies dramatically, from 1.0K tokens for Level 1 to 7.5K for Level 5 (\cref{fig:appendix_math500}b). The marginal utility curve (\cref{fig:appendix_math500}c) clearly shows earlier diminishing returns for easier problems.   

\subsubsection{Case Studies of Negative Flips}
\label{sec:appendix_cases}

We manually examined 80 randomly sampled negative flip cases from R1-32B on AIME and categorized them into three types (\cref{tab:qualitative}). Below we provide representative examples from each category.

\begin{table}[h]
\centering
\small
\begin{tabular}{lcc}
\hline
\textbf{Category} & \textbf{Count} & \textbf{Percentage} \\
\hline
(A) Genuine overthinking & 54 & 67.5\% \\
(B) Exploration divergence & 16 & 20.0\% \\
(C) Degradation artifacts & 10 & 12.5\% \\
\hline
\end{tabular}
\caption{\textbf{Qualitative analysis of negative flips.} Most negative flips (67.5\%) involve genuine overthinking where models explicitly abandon correct answers.}
\label{tab:qualitative}
\end{table}

\paragraph{Category A: Genuine Overthinking}
\textit{Problem:} AIME 2024 Problem 7 (combinatorics).

At 4K tokens, the model correctly identifies the answer as 220 using a standard counting argument. At 8K tokens, the model revisits the problem: ``\textit{Wait, I should double-check by considering an alternative approach... Actually, I think I may have overcounted. Let me reconsider the boundary cases...}'' The model then incorrectly adjusts its count to 198, second-guessing the correct initial solution.

This pattern, where explicit reconsideration leads to abandoning correct answers, accounts for 67.5\% of negative flips.

\paragraph{Category B: Exploration Divergence}
\textit{Problem:} AIME 2025 Problem 3 (number theory).

At 3K tokens, the model solves the problem correctly using modular arithmetic. At 7K tokens, the model attempts a different approach: ``\textit{Let me try solving this using the Chinese Remainder Theorem instead...}'' While the alternative approach is mathematically valid, the model makes an arithmetic error in the execution, arriving at an incorrect answer.

This category (20\%) represents cases where extended exploration finds valid alternative methods but introduces execution errors.

\paragraph{Category C: Degradation Artifacts}
\textit{Problem:} AIME 2024 Problem 12 (geometry).

At 5K tokens, the model provides a correct answer. At 12K tokens, the reasoning becomes increasingly repetitive and unfocused, with the model restating the same equations multiple times without progress. The final answer differs from the correct one without clear justification.

This category (12.5\%) represents cases where extended generation leads to output degradation without explicit reasoning errors.

\begin{figure*}[t!]
  \centering
  \includegraphics[width=0.85\linewidth]{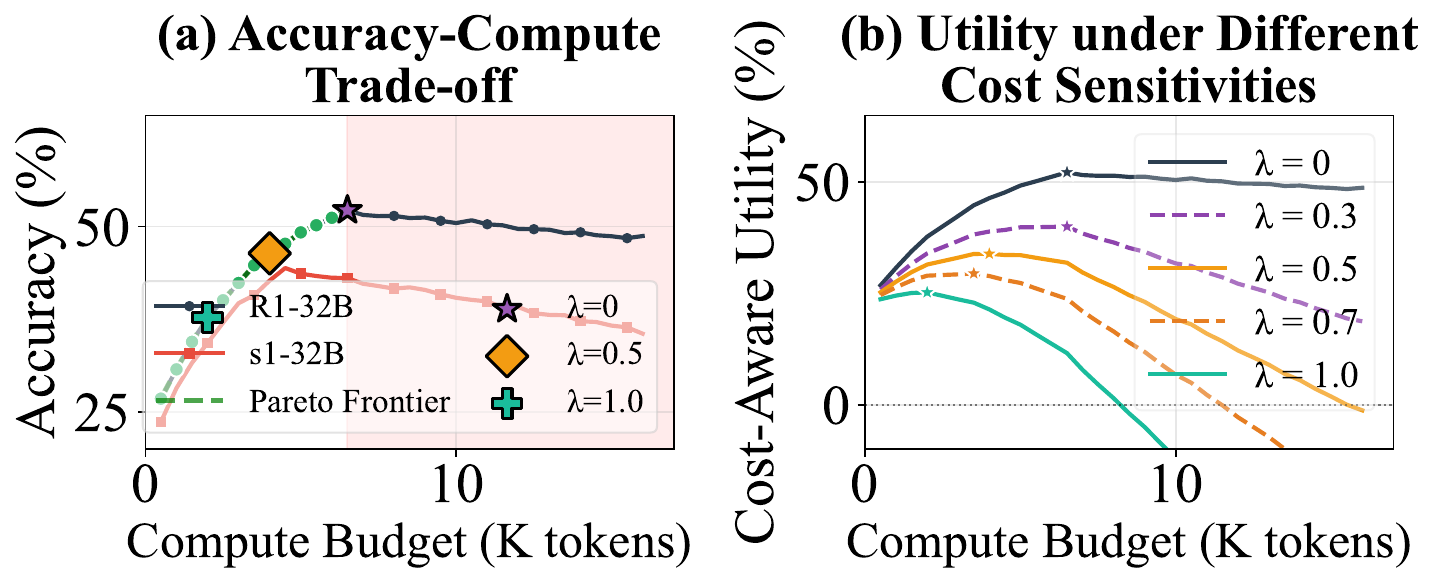}
  \caption{\textbf{Cost-aware evaluation reveals optimal stopping points.} \textit{(a)} The Pareto frontier shows the accuracy-compute trade-off. Markers indicate optimal budgets under different $\lambda$ values: at $\lambda{=}0$ (cost-agnostic), peak accuracy budget is optimal (not maximum, due to overthinking); at $\lambda{=}1.0$ (cost-sensitive), early stopping achieves higher utility. \textit{(b)} Utility curves shift as cost sensitivity increases, with optimal stopping points moving leftward.
  }
  \label{fig:efficiency}
\end{figure*}

\begin{figure*}[t!]
  \centering
  \includegraphics[width=0.9\linewidth]{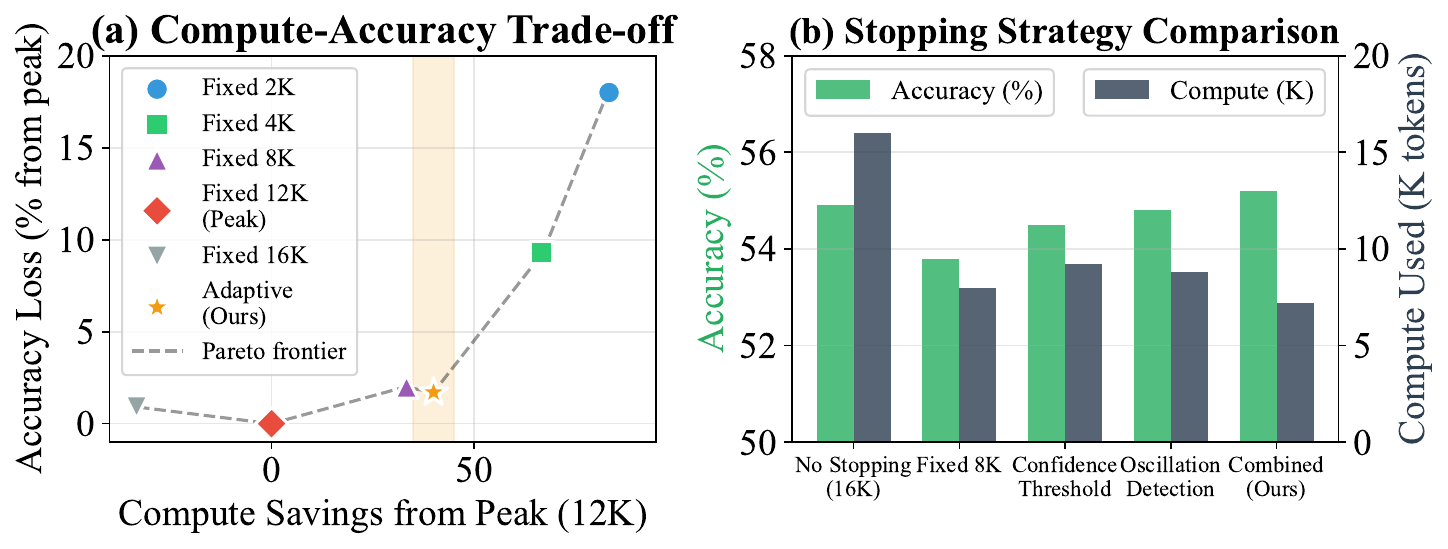}
  \caption{\textbf{Early Stopping Validation.} (a) The compute-accuracy trade-off for different stopping constraints. (b) Strategy comparison showing our combined approach achieves strong accuracy with significant compute savings.}
  \label{fig:appendix_stopping}
\end{figure*}

\section{Cost-Aware Evaluation}
\label{sec:cost}

\mysubsection{Motivation}
\label{sec:cost_motivation}

Current evaluations of test-time scaling report accuracy at various compute budgets, implicitly treating computation as free. In practice, inference cost is a primary deployment concern: generating 16,000 tokens costs 32$\times$ more than generating 500 tokens. Our findings in \cref{sec:experiments} reveal that much computation at high budgets provides minimal benefit or actively harms performance through overthinking. This motivates evaluation frameworks that capture the accuracy-compute trade-off.

Just as \citet{jurayj2025final} extended test-time scaling evaluation by introducing risk-aware utility functions, we propose \textit{cost-aware} metrics that penalize excessive computation. Where their work asks ``should the model answer at all?'', we ask ``how long should the model think?''

\mysubsection{Efficiency Metrics}
\label{sec:metrics}

We define a cost-aware utility function that balances accuracy against compute:
\begin{equation}
    U_\lambda(t) = \text{Acc}(t) - \lambda \cdot \frac{t}{t_{\max}}
\end{equation}
where $\text{Acc}(t) \in [0, 1]$ is accuracy at budget $t$, $t_{\max}$ is the maximum budget evaluated, and $\lambda \geq 0$ controls cost sensitivity. We consider three evaluation scenarios analogous to the risk levels in selective question answering:\\
 \textbf{Cost-Agnostic} ($\lambda = 0$): Maximize accuracy regardless of compute. This is the standard evaluation paradigm.\\
 \textbf{Cost-Balanced} ($\lambda = 0.5$): Accuracy gains must justify compute expenditure. A 1\% accuracy improvement requires $\leq$2\% additional compute.\\
\textbf{Cost-Sensitive} ($\lambda = 1.0$): Strong efficiency preference. Only compute that yields proportional accuracy gains is justified.

\mysubsection{Main Results}
\label{sec:stopping}


Under cost-agnostic evaluation ($\lambda{=}0$), the optimal strategy is to use compute up to peak accuracy. As $\lambda$ increases, optimal budgets shift dramatically lower. At $\lambda{=}0.5$, stopping at $\sim$6K tokens yields $\sim$50\% compute reduction with only $\sim$6\% accuracy loss, while $\lambda{=}1.0$ favors $\sim$2K tokens (\cref{fig:efficiency}).
We further validate that indicator-based early stopping can achieve 97\% of peak accuracy while using only 60\% of compute (\cref{sec:appendix_stopping}).

\subsection{Early Stopping Validation}
\label{sec:appendix_stopping}

\cref{fig:appendix_stopping} validates our early-stopping approach. (a) shows the compute-accuracy trade-off, demonstrating how accuracy changes with varying compute limits. (b) compares the performance of different stopping strategies on AIME: our combined indicator-based approach effectively reduces compute while maintaining competitive accuracy compared to fixed token limits.

\section{Conclusion}
We analyze diminishing returns in test-time compute scaling, finding that (1) marginal utility decreases substantially at high budgets, and (2) models exhibit ``overthinking,'' abandoning correct answers after extended reasoning. We introduce flip event tracking and cost-aware evaluation metrics to capture accuracy-compute trade-offs. We encourage the community to report efficiency frontiers alongside accuracy curves.

\section*{Limitations}

Our analysis focuses on mathematical and scientific reasoning tasks; overthinking may manifest differently in other domains. While our validation experiments confirm that overthinking occurs in natural model behavior (not just forced continuations), more naturalistic approaches could strengthen these findings. We evaluate only open-weight models; proprietary systems may exhibit different patterns. Additionally, while our qualitative analysis suggests genuine reconsideration behavior in 67.5\% of negative flips, establishing definitive causal mechanisms underlying overthinking requires further investigation through controlled interventions.

\section*{Ethics Statement}

This work analyzes the computational efficiency of large language model reasoning, which we believe has positive ethical implications. By identifying overthinking behaviors where extended computation degrades performance, our findings can help reduce unnecessary energy consumption and carbon emissions associated with LLM inference.

\section*{Acknowledgements}
This work is supported by National Natural Science Foundation of China (Grant No. 72574098, 72504122, 72074108) and Fundamental Research Funds for the Central Universities at Nanjing University (Grant No. 010814370338), Jiangsu Young Talents in Social Sciences and Tang Scholar of Nanjing University.

\bibliography{custom}
\newpage
\appendix

\onecolumn

\section{Natural Long Reasoning Analysis}
\label{sec:appendix_natural}

This section provides detailed analysis supporting the validation experiment in \cref{sec:natural}.

\paragraph{Sample Selection} We identify natural long-reasoning samples by running R1-32B on all problems \textit{without} budget forcing, allowing the model to conclude naturally. From 560 total samples (AIME + MATH-500), we find 312 samples (55.7\%) where the model naturally generated $>$8K tokens. These samples tend to be harder problems (78\% are Level 4-5 on MATH-500 difficulty scale).

\paragraph{Accuracy by Natural Length} \cref{tab:natural_detail} shows accuracy stratified by the model's natural output length. Interestingly, problems where the model naturally writes more tokens tend to have lower accuracy, suggesting that the model's own length choice correlates with problem difficulty and uncertainty.

\begin{table}[h]
\centering
\small
\begin{tabular}{lcc}
\hline
\textbf{Natural Length} & \textbf{N} & \textbf{Accuracy} \\
\hline
$<$4K tokens & 89 & 71.9\% \\
4--8K tokens & 159 & 58.5\% \\
8--12K tokens & 198 & 51.0\% \\
$>$12K tokens & 114 & 44.7\% \\
\hline
\end{tabular}
\caption{\textbf{Accuracy by natural output length.} Longer natural outputs correlate with lower accuracy, suggesting the model writes more when uncertain.}
\label{tab:natural_detail}
\end{table}

\paragraph{Second-Guessing Behavior} Among the 312 natural long-reasoning samples, we identified instances where the model explicitly reconsiders its answer using pattern matching for phrases like ``wait'', ``actually'', ``let me reconsider'', ``I made a mistake'', etc. We find that 71\% (221/312) of these samples contain at least one explicit reconsideration, and samples with reconsideration have 12\% lower accuracy than those without, providing further evidence for the overthinking hypothesis.

\end{document}